\documentclass[conference]{IEEEtran}
\usepackage{amsmath}
\usepackage{amsfonts}
\usepackage{multirow}
\usepackage{graphicx}
\usepackage{footnote}
\usepackage{times}
\usepackage{textcomp}
\makesavenoteenv{tabular}
\makesavenoteenv{table}

\ifCLASSINFOpdf
\else
\fi
\begin{document}
%
\title{Large-Scale Detection of Non-Technical Losses in Imbalanced Data Sets}



%
\author{\IEEEauthorblockN{Patrick Glauner\IEEEauthorrefmark{1},
Andre Boechat\IEEEauthorrefmark{1},
Lautaro Dolberg\IEEEauthorrefmark{1}, 
Radu State\IEEEauthorrefmark{1},
Franck Bettinger\IEEEauthorrefmark{2},
Yves Rangoni\IEEEauthorrefmark{2} \\and
Diogo Duarte\IEEEauthorrefmark{2}}
\IEEEauthorblockA{\IEEEauthorrefmark{1}Interdisciplinary Centre for Security, Reliability and Trust, University of Luxembourg\\
2721 Luxembourg, Luxembourg\\
Email: \{first.last\}@uni.lu}
\IEEEauthorblockA{\IEEEauthorrefmark{2}CHOICE Technologies Holding S\`arl\\
2-4, rue Eug\`ene Ruppert \\
2453 Luxembourg, Luxembourg \\
Email: \{first.last\}@choiceholding.com}
}


\maketitle

\begin{abstract}
Non-technical losses (NTL) such as electricity theft cause significant harm to our economies, as in some countries they may range up to 40\% of the total electricity distributed. Detecting NTLs requires costly on-site inspections. Accurate prediction of NTLs for customers using machine learning is therefore crucial. To date, related research largely ignore that the two classes of regular and non-regular customers are highly imbalanced, that NTL proportions may change and mostly consider small data sets, often not allowing to deploy the results in production. In this paper, we present a comprehensive approach to assess three NTL detection models for different NTL proportions in large real world data sets of 100Ks of customers: Boolean rules, fuzzy logic and Support Vector Machine. This work has resulted in appreciable results that are about to be deployed in a leading industry solution. We believe that the considerations and observations made in this contribution are necessary for future smart meter research in order to report their effectiveness on imbalanced and large real world data sets.
\end{abstract}

\begin{IEEEkeywords}
Electricity Theft Detection, Fuzzy Logic, Imbalanced Classification, Non-Technical Losses, Support Vector Machine
\end{IEEEkeywords}

%
\IEEEpeerreviewmaketitle

\section{Introduction}
Electrical power grids are the backbone of today's society. Losses during generation and distribution cause major problems, including financial losses to electricity providers and a decrease of stability and reliability.
They can be classified into technical losses and non-technical losses. Technical losses are naturally occurring and mainly include losses to power dissipation in electrical components, such as in generators, transformers and transmission lines due to internal electrical resistance. They are possible to detect and control given a knowledge of the quantities of loads.

Non-technical losses (NTL) faced by electricity providers include, but are not limited to, electricity theft by rewiring or manipulating meters. Other types include faulty meters and errors in meter readings and billing. There are different estimates of the financial losses caused by NTLs and they can range up to 40\% of the total electricity distributed in countries such as Brazil, India, Malaysia or Lebanon \cite{rules_high}, \cite{nagi_fuzzy}. They are also of relevance in developed countries, for example estimates of NTLs in the US range from USD 1-6 billion \cite{rules_high}.

In order to detect NTLs, inspections of customers are carried out, based on predictions whether there may be a NTL at a customer. The inspection results are then used in the learning of algorithms in order to improve predictions.
However, carrying out inspections is expensive, as it requires physical presence of technicians. It is therefore important to make accurate predictions in order to reduce the number of false positives.

Detecting NTLs is challenging because of the wide range of possible causes of NTLs, such as different fraudulent types of customers. From a machine learning perspective, a key problem is the imbalance of the data, meaning that there are significantly more regular customers than customers with NTLs.
We believe that this property has not adequately been addressed and reported in the literature. We therefore assess various prediction models for different proportions of NTLs in the data and discuss representative performance measures for a reliable assessment of them.
We believe that an accurate discussion of this topic is necessary for future work on NTL detection in a smart meter environment.

The rest of this paper is organized as follows. Section \ref{chapter:review} provides a literature review of NTL detection and its challenges. Section \ref{chapter:ntl} describes different proposed NTL detection models and the respective data set. Section \ref{chapter:eval} presents experimental results and comparison of the models on the data for different NTL proportions in the data. Section \ref{chapter:end} summarizes this work and provides an outreach on future work.

\section{Related work}
\label{chapter:review}
\subsection{Literature review}
NTL detection can be treated as a special case of fraud detection, for which a general survey is provided in \cite{fraud_survey}. It highlights two approaches as key methods to detect fraudulent behavior in credit card fraud, computer intrusion and telecommunications fraud: (i) expert systems that represent domain knowledge in order to make decisions typically using hand-crafted rules and (ii) data mining or machine learning techniques that employ statistics to learn patterns from sample data in order to make decisions for future unseen data. Both approaches have their justification and neither is generally better or worse than the other one in artificial intelligence \cite{hand_written}. 

One method to detect NTLs is to calculate the energy balance \cite{energy_balance}, which requires topological information of the network. This does not work accurately for those reasons: (i) in developing countries, network topology undergoes continuous changes in order to satisfy the rapidly growing demand of electricity, (ii) infrastructure may break and lead to wrong energy balance calculations and (iii) it requires transformers, feeders and connected meters to be read at the same time.

Another approach is to analyze the customer load profile using artificial intelligence methods, such as machine learning or expert systems.
Support Vector Machines (SVM) are used in \cite{nagi}, working on daily average consumption features of the last 24 months for less than 400 highly imbalanced training examples, ignoring the class imbalance in the results reported. That work is combined with fuzzy logic \cite{nagi_fuzzy} or genetic algorithms \cite{nagi_genetic}, focusing on an optimization of the SVM output.
A rule-based expert system system outperforms a SVM in \cite{rules_high} for an unknown amount of customers, focusing on high performance implementations.
Fuzzy logic following C-means fuzzy clustering is applied to a data set of \texttildelow 20K customers in \cite{angelos_fuzzy}.
Furthermore, neural networks using hand-crafted features calculated from the consumption time series  plus customer-specific pre-computed attributes are used in \cite{nn1} for \texttildelow 1K balanced customers. Applying smart half-hour meter readings of three weeks of \texttildelow 6K customers are fed into a neural network in \cite{nn2}.
Optimum-path forest are applied to NTL detection in \cite{path_forest} for \texttildelow 10K customers outperforming different SVMs and a neural network.
A different method is to estimate NTLs by subtracting an estimate of the technical losses from the overall losses \cite{sahoo_technical_loss}. In many electricity grids it may be challenging to accumulate the entire losses and furthermore, this method does not scale to large numbers of meters.
The class imbalance problem of electricity theft detection has initially been addressed in \cite{imbalanced_opponent}. It applies an ensemble of two SMVs, optimum-path forest and C4.5 decision tree learning to \texttildelow 300 on-field inspection test data. However, the degree of imbalance of the \texttildelow 1.5K training examples is not reported. Furthermore, in the optimization of the classifiers, the true negative rate is ignored, which results in too many costly inspections of non-fraudulent customers.


Many of these results are constrained to either just a few test examples or report accuracies on highly imbalanced classes.
To date, working on large and long-term data sets and properly measuring the performance of the classifiers on imbalanced data sets has not adequately been studied in the literature. However, ignoring the class imbalance in reported results is also true for many other machine learning applications.
In this paper, we focus on large data sets comprising each of \texttildelow 100K inspection results spanning four years of consumption data and apply different NTL detection methods on it. We particularly address the class imbalance problem using accurate performance measures.

\subsection{Challenge of supervised learning for anomaly detection}
It must be noted that most NTL detection methods are supervised. Anomaly detection - a superclass of NTL - is generally challenging to learn in a supervised manner for the reasons stated in \cite{ng}:
(i) anomaly data sets contain a very small number of positive examples and large number of negative examples, resulting in imbalanced classes, (ii) it is used for many different kinds of anomalies as it is hard for any algorithm to learn from just a few positive examples what the anomalies might look like and (iii) there may be also future anomalies which may look completely different to any of the anomalous examples learned so far.
In contrast, supervised learning works best for (i) large numbers of both positive and negative examples, (ii) when there are enough positive examples so that the algorithm can get a sense of what positive examples might look like and (iii) future positive examples are likely to be similar to the ones in the training set.

\section{NTL detection}
\label{chapter:ntl}

\subsection{Data}
\label{chapter:ntl:data}
The data used in this paper is from an electricity provider in Brazil. It consists of three parts: (i) \texttildelow 700K customer data, such as location, type, etc., (ii) \texttildelow 31M monthly consumption data from January 2011 to January 2015 such as consumption in kWh, date of meter reading and number of days between meter readings and (iii) \texttildelow 400K inspection data such as presence of fraud or irregularity, type of NTL and inspection notes.

Most inspections do not find NTLs, making the classes highly imbalanced. In order for the models to be applied to other regions or countries, they must be assessed on different NTL proportions. Therefore, the data was subsampled using 17 different NTL proportion levels: 0\%, 0.1\%, 1\%, 2\%, 3\%, 4\%, 5\%, 10\%, 20\%, 30\%, 40\%, 50\%, 60\%, 70\%, 80\%, 90\% and 100\%. Each sample contains \texttildelow 100K inspection results.

\subsection{Models}
In this Section, the different models for NTL detection of this paper are described. The first model is a CHOICE Technologies product based on Boolean logic and is used as a baseline. It is extended to fuzzy logic in the second model in order to smoothen the decision making process. The third model is a Support Vector Machine, a state-of-the-art machine learning algorithm.

\subsubsection{Boolean logic}
This model is an expert system, it consists of hand-crafted rules created by the CHOICE Technologies expert team which are conjunctions of (in)equality terms, such as:
\begin{align}
(N_1 > v_1) \wedge (N_1 < v_2) \wedge (N_2 < v_3) \wedge (N_3 = v_4) ...
\end{align}
$N_x$ is a so-called attribute. Possible attributes are change of consumption over the last 3 months, slope of consumption curves, etc. and $v_x$ is a numeric value. In total, 42 attributes are used in 14 rules.
If at least one rule outcome is true, that customer is considered to potentially cause a NTL.

\subsubsection{Fuzzy logic}
Fuzzy systems \cite{fuzzy} have a long tradition in control applications allowing to implement expert knowledge in a softer decision making process. They allow to relate to classes of objects, breaking up boundaries, making membership a matter of degree. In this paper, the 14 Boolean rules were fuzzified and incorporated in a Mamdani fuzzy system using the centroid defuzzification method \cite{fuzzy}. Fuzzy rules rely on membership functions.
The number of membership functions for each attribute depends on the ranges of values found in the rules among which 1 attribute has 1 function, 32 attributes have 2 membership functions and 9 attributes have 4 functions. In most cases, trapezoid membership functions are used to keep the model simple. The exact parameters, such as membership function boundaries or the mean of sigmoid membership functions were determined from the distribution of attribute values.

However, these parameters could be optimized using: (i) gradient techniques \cite{frbs}, (ii) genetic algorithms \cite{frbs} or (iii) neuro-fuzzy systems \cite{neuro_fuzzy}. Techniques (i) and (ii) are highly constrained optimization problems due to dependence among parameter values to keep the fuzzy system valid.
Technique (i) was studied further and its results are reported in Section~\ref{chapter:eval}.

\subsubsection{Support Vector Machine}
A Support Vector Machine (SVM) \cite{vapnik} is a maximum margin classifier, i.e. it creates a maximum separation between classes. Therefore, a SVM is less prone to overfitting than other classifiers, such as a neural network \cite{svm_overfit}. Support vectors hold up the separating hyperplane. In practice, they are just a small fraction of the training examples.

The training of a SVM can be defined as a Lagrangian dual problem having a convex cost function. In that form, the optimization formulation is written in terms of only the dot product $x^{(i)}\cdot x^{(j)}$ between points in the input space.
By default, the separating hyperplane is linear.
For complex problems, it is advantageous to map the data set to a higher dimension space, where it is possible to separate them using a linear hyperplane. A kernel is an efficient function that implicitly computes the dot product in the higher dimensional space.
A popular kernel is the Gaussian radial basis function: $K(u,v) = \exp(-\gamma \lVert u -v\rVert^2)$.

Inspired by \cite{nagi}, for $M$ customers $\{0, 1, ..., M - 1\}$ over the last $N$ months $\{0, 1, ..., N -1\}$, a feature matrix $F$ is computed, in which element $F_{m, d}$ is a daily average kWh consumption feature during that month:
\begin{align}
x_d^{(m)} = \frac{L_d^{(m)}}{R^{(m)}_{d} - R^{(m)}_{d-1}}
\end{align}
where for customer $m$, $L_d^{(m)}$ is the kWh consumption increase between the meter reading to date $R^{(m)}_d$ and the previous one $R^{(m)}_{d-1}$. $R^{(m)}_{d} - R^{(m)}_{d-1}$ is the number of days between both meter readings of customer $m$.
Similarly, a binary target vector $T$ is created in which element $T^{(m)}$ is the most recent inspection result for customer $m$ in the respective period of time.
NTLs are encoded by 1 if they are detected and 0 if not.

\section{Evaluation}
\label{chapter:eval}

\subsection{Metrics}
In many classification problems, the classification rate, or accuracy is used as a performance measure. Given the number of true positives (TP), true negatives (TN), false positives (FP) and false negatives (FN): $ACC = \frac{TP+TN}{TP + TN + FP + FN}$.
However, many publications ignore that it is only of minor expressiveness for imbalanced classes. For a NTL detection example, given a data set of 990 negative and 10 positive test examples, a classifier that always predicts negative has an accuracy of $0.99$. This example clearly demonstrates that other performance measures must be used for NTL detection.
%
The recall is a measure of the proportion of the true positives found. It is also named true positive rate (TPR) or sensitivity: $Recall = \frac{TP}{TP + FN}$.
%
The specificity is a measure of the proportion of the true negatives classified as negative. It is also named true negative rate (TNR): $Specificity = \frac{TN}{TN + FP}$.
The false positive rate (FPR) is $1 - TNR$.
A receiver operating characteristic (ROC) curve plots the TPR against the FNR. The area under the curve (AUC) is a performance measure between 0 and 1, where any binary classifier with an AUC $> 0.5$ performs better than random guessing.
While in many applications multiple thresholds are used to generate points plotted in a ROC curve, the AUC can also be computed for a single point, when connecting it with straight lines to $(0, 0)$ and $(1, 1)$ as shown in \cite{area_roc}: $AUC = \frac{Recall + Specificity}{2}$.

For NTL detection, the goal is to reduce the FPR  to decrease the number of costly inspections, while increasing the TPR to find as many NTL occurrences as possible. In order to assess a NTL prediction model using a single performance measure, the AUC is the most suitable.

\subsection{Methodology}
Throughout the experiments, consumption readings and inspection result data are used. Further data, such as location of customers are not used.
In the comparison of the three classifiers, the AUC performance measure is used for the different levels of NTL proportion mentioned in Section~\ref{chapter:ntl:data}.
We assessed different values for the number of the most recent meter readings $N$. Only customers with complete time series of the last $N$ months before the respective inspection are considered. The larger $N$, the less data is available. At least 12 months should be considered in order to represent seasonality effects. Experiments for the last 12, 18 and 24 months were carried out, for which 12 months have proven to lead to the best results as the other experiments lead to more overfitting. Due to lack of space, those results are omitted.

The SVM is the only classifier that requires training in our experiments. However, since it is a binary classifier, it could not be trained on NTL proportions of 0\% and 100\%. For the NTL proportions used for training, 10-folded cross validation is performed for every NTL proportion, splitting the data into a 60\%/20\%/20\% training/validation/test ratio. The AUC score is used as the validation measure to pick the best classifier fold. Throughout the experiments, a linear SVM is used. The same experiments were repeated using a Gaussian Kernel, which proved to overfit for all NTL proportions.

\subsection{Implementation details}
The Boolean and fuzzy classifiers were implemented in \texttt{MATLAB}, the latter using the Fuzzy Logic Toolbox \cite{matlab}. The SVM classifier was implemented in Python using \texttt{scikit-learn} \cite{scikit}, which builds on top of \texttt{LIBSVM} \cite{libsvm}. The regularization parameter and the inverse variance parameter $\gamma$ of the Gaussian kernel were not optimized explicitly, as \texttt{scikit-learn} optimizes them automatically.
Using 10-fold cross-validation to train 10 SVMs and to select the best one takes about 2 minutes per NTL proportion on a state-of-the-art i5 notebook. Using the Boolean or fuzzy systems to classify the same amount of data takes about 1 second. However, both classifiers use pre-computed customer-specific attributes. Computing them takes a couple of hours in a cloud infrastructure.

\subsection{Comparison of classifier performance}

\begin{figure}[!t]
\centering
\includegraphics[width=0.5\textwidth]{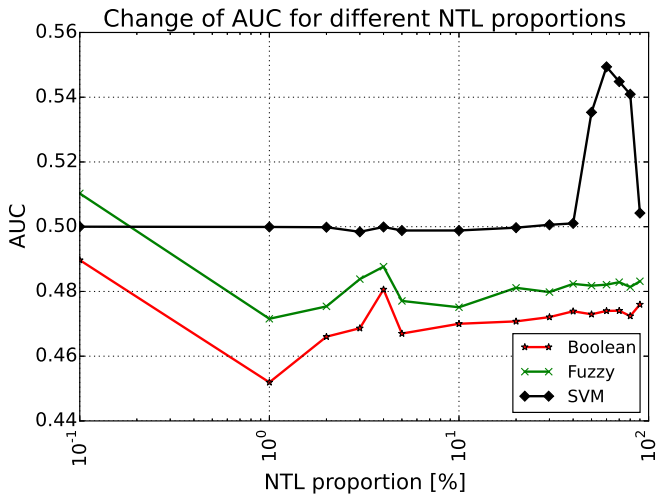}
\caption[XXX]{Comparison of classifiers tested on different NTL proportions.}
\label{fig:AUC1}
\end{figure}

For different NTL proportions, the change of test AUC for the Boolean and fuzzy systems and the SVM can be observed in Fig.~\ref{fig:AUC1}.
The Boolean classifier has an AUC $< 0.5$ for all NTL proportions and therefore performs worse than random guessing. The same applies for the fuzzy system, except for a NTL proportion of 0.1\%.
The SVM performs only (noticeably) better than random guessing for NTL proportions between 50\% and 80\%.

\begin{figure}[!t]
\centering
\includegraphics[width=0.5\textwidth]{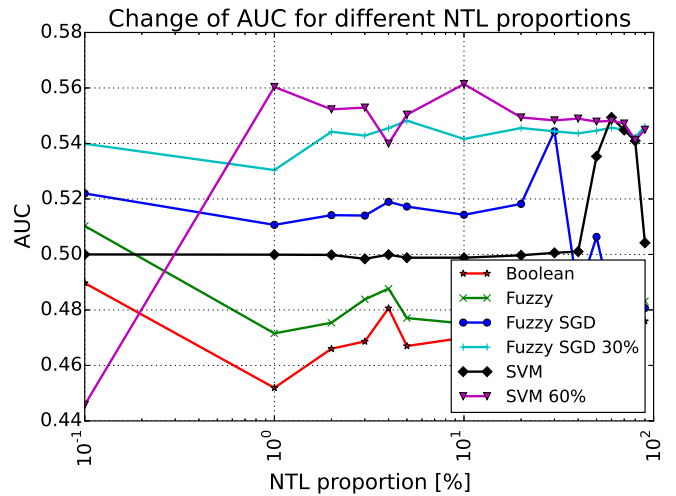}
\caption{Comparison of optimized classifiers tested on different NTL proportions.}
\label{fig:AUC2}
\end{figure}

Given the theory of fuzzy systems and their potential, the parameters of the fuzzy system were optimized using stochastic gradient descent (SGD) for each of the 15 binary NTL proportions: 0.1\% to 90\%. Out of the 15 optimized fuzzy systems, the one with the greatest AUC test score is picked and tested on all NTL proportions. The fuzzy system trained on 30\% and tested on all NTL proportions - Fuzzy SGD 30\% - significantly outperforms both, the Boolean and fuzzy systems, as shown in Fig.~\ref{fig:AUC2}.

%

The same methodology as for the optimized fuzzy system is applied to the SVM. SVMs are trained on all 15 binary NTL proportions and SVM 60\%, the SVM trained on 60\% NTL proportion, is selected because of the greatest AUC test score. Its performance compared to the Boolean and Fuzzy SGD 30\% are shown in Fig.~\ref{fig:AUC2}. In summary, SVM 60\% performs in a similar range as Fuzzy SGD 30\% compared to the Boolean system, except for very small NTL proportions $< 1\%$.

However, comparing the confusion matrices of both classifiers, they perform very differently as shown in Tables~\ref{table:compared5} and \ref{table:compared20} for selected NTL levels of 5\% and 20\%, respectively.
The optimized fuzzy system has a higher TNR, but lower TPR compared to the optimized SVM. In return, the SVM has a higher TPR, but a lower FNR.

\begin{table}[!t]
\renewcommand{\arraystretch}{1.3}
\caption{Normalized conf. matrices for test on 5\% NTL proportion.}
\label{table:compared5}
\centering
\begin{tabular}{cc|c|c|c|}
\cline{3-4}
&  & \multicolumn{2}{c|}{Predicted} \\ \cline{1-4}
\multicolumn{1}{ |c  }{\multirow{2}{*}{Classifier} } &
\multicolumn{1}{ |c| }{\multirow{2}{*}{Actual} } &  TNR & FPR    \\ \cline{3-4}
\multicolumn{1}{ |c  }{}                        &
\multicolumn{1}{ |c| }{} & FNR & TPR   \\ \cline{1-4}
\multicolumn{1}{ |c  }{\multirow{2}{*}{Boolean} } &
\multicolumn{1}{ |c| }{\multirow{2}{*}{Actual} } & 0.53 & 0.47   \\ \cline{3-4}
\multicolumn{1}{ |c  }{}                        &
\multicolumn{1}{ |c| }{} & 0.60  & 0.40  \\ \cline{1-4}
\multicolumn{1}{ |c  }{\multirow{2}{*}{Fuzzy SGD 30\%} } &
\multicolumn{1}{ |c| }{\multirow{2}{*}{Actual} } & 0.87 & 0.13   \\ \cline{3-4}
\multicolumn{1}{ |c  }{}                        &
\multicolumn{1}{ |c| }{} &  0.77  & 0.23    \\ \cline{1-4}
\multicolumn{1}{ |c  }{\multirow{2}{*}{SVM 60\%.} } &
\multicolumn{1}{ |c| }{\multirow{2}{*}{Actual} }  & 0.36  & 0.64     \\ \cline{3-4}
\multicolumn{1}{ |c  }{}                        &
\multicolumn{1}{ |c| }{} & 0.26 & 0.74   \\ \cline{1-4}
\end{tabular}
\end{table}

\begin{table}[!t]
\renewcommand{\arraystretch}{1.3}
\caption{Normalized conf. matrices for test on 20\% NTL proportion.}
\label{table:compared20}
\centering
\begin{tabular}{cc|c|c|c|}
\cline{3-4}
&  & \multicolumn{2}{c|}{Predicted} \\ \cline{1-4}
\multicolumn{1}{ |c  }{\multirow{2}{*}{Classifier} } &
\multicolumn{1}{ |c| }{\multirow{2}{*}{Actual} } &  TNR & FPR    \\ \cline{3-4}
\multicolumn{1}{ |c  }{}                        &
\multicolumn{1}{ |c| }{} & FNR & TPR   \\ \cline{1-4}
\multicolumn{1}{ |c  }{\multirow{2}{*}{Boolean} } &
\multicolumn{1}{ |c| }{\multirow{2}{*}{Actual} } & 0.53 & 0.47   \\ \cline{3-4}
\multicolumn{1}{ |c  }{}                        &
\multicolumn{1}{ |c| }{} &  0.58  &  0.42 \\ \cline{1-4}
\multicolumn{1}{ |c  }{\multirow{2}{*}{Fuzzy SGD 30\%} } &
\multicolumn{1}{ |c| }{\multirow{2}{*}{Actual} } & 0.87 & 0.13   \\ \cline{3-4}
\multicolumn{1}{ |c  }{}                        &
\multicolumn{1}{ |c| }{} & 0.78  & 0.22   \\ \cline{1-4}
\multicolumn{1}{ |c  }{\multirow{2}{*}{SVM 60\%} } &
\multicolumn{1}{ |c| }{\multirow{2}{*}{Actual} } & 0.35 & 0.65    \\ \cline{3-4}
\multicolumn{1}{ |c  }{}                        &
\multicolumn{1}{ |c| }{} &  0.25 & 0.75   \\ \cline{1-4}
\end{tabular}
\end{table}

\subsection{Discussion}
The initial Boolean and fuzzy models perform worse than random guessing and are therefore not suitable for real data, as they trigger too many inspections while not many of them will lead to NTL detection.
Optimized fuzzy and SVM models trained on 30\% and 60\% NTL proportion, respectively, result in significantly greater AUC scores. However, both perform very differently, as the optimized fuzzy system is more conservative in NTL production. In contrast, the optimized SVM is more optimistic, leading also to a higher FPR.
In general, neither can be named better than the other one, as picking the appropriate model from these two is subject to business decisions.

However, this work also demonstrates that for real data, NTL classifiers using only the consumption profile are limited.
Therefore, it is desirable to use more features like location, inspection notes, etc.
Another issue with the real data is the potential bias of inspections so that this sample of customers does not represent the overall population of customers. We expect a correction of the bias to lead to better predictions, too.

\section{Conclusion and future work}
\label{chapter:end}
In this work, we have proposed three models for NTL detection for large data sets of 100K customers: Boolean, fuzzy and Support Vector Machine.
In contrast to other results reported in the literature, the optimized fuzzy and SVM models were assessed for varying NTL proportions on imbalanced real world consumption data. 
Both have an AUC $> 0.5$ for all NTL proportions $> 0.1\%$ and significantly outperform simple Boolean or unoptimized fuzzy models. 
The improved models are about to be deployed in a CHOICE Technologies product.
The contribution methodology is necessary for future smart meter research, in order to report their effectiveness in imbalanced and large real world data sets.

We are planning to evaluate unsupervised methods, in particular deep learning, in order to detect NTL more accurately by finding hidden correlations in the data. Furthermore, we are planning to use other features in our models, such as the location and latent features and to investigate cost-based optimization in order to maximize the total electricity recovered through inspections.
Also, we are planning to make our implementations faster and more scalable using Apache Spark \cite{spark}.





%

\end{document}